\documentclass{article}
\usepackage{howtowrite}

\usepackage{times}
\usepackage{soul}
\usepackage{url}
\usepackage[hidelinks]{hyperref}
\usepackage[utf8]{inputenc}
\usepackage[small]{caption}
\usepackage{graphicx}
\usepackage{amsmath}
\usepackage{booktabs}
\usepackage{algorithm}
\usepackage{algorithmic}
\usepackage{multirow}
\usepackage{verbatim}
\usepackage{CJKutf8}  
\usepackage{subfigure}
\usepackage{mhchem}

\title{How to Write High-quality News on Social Network? Predicting News Quality by Mining Writing Style}

\author{
Yuting Yang$^{1,2}$
\and
Juan Cao$^{1,2}$\and
Mingyan Lu$^{1,2}$\and
Jintao Li$^{1}$\And
Chia-Wen Lin$^3$
\affiliations
$^{1}$Key Laboratory of Intelligent Information Processing  Center for Advanced Computing Research,
Institute of Computing Technology, CAS, Beijing, China\\
$^{2}$University of Chinese Academy of Sciences, Beijing, China\\
$^3$Department of Electrical Engineering, National Tsing Hua University
\emails
\{yangyuting, caojuan, lumingyan, jtli\}@ict.ac.cn,
cwlin@ee.nthu.edu.tw
}
\begin{document}
\begin{CJK}{UTF8}{gkai} 
\maketitle

\begin{abstract}
Rapid development of Internet technologies promotes traditional newspapers to report news on social networks. However, people on social networks may have different needs which naturally arises the question: whether can we analyze the influence of writing style on news quality automatically and assist writers in improving news quality? It's challenging due to writing style and `quality'  are hard to measure. First, we use `popularity' as the measure of `quality'. It is natural on social networks but brings new problems: popularity are also influenced by event and publisher. So we design two methods to alleviate their influence. Then, we proposed eight types of linguistic features (53 features in all) according eight writing guidelines and analyze their relationship with news quality. The experimental results show these linguistic features influence greatly on news quality. Based on it, we design a news quality assessment model on social network (SNQAM). SNQAM performs excellently on predicting quality, presenting interpretable quality score and giving accessible suggestions on how to improve it according to writing guidelines we referred to.

\end{abstract}

\section{Introduction}
\begin{figure*}[h]
	\centering
	\includegraphics[scale=0.5]{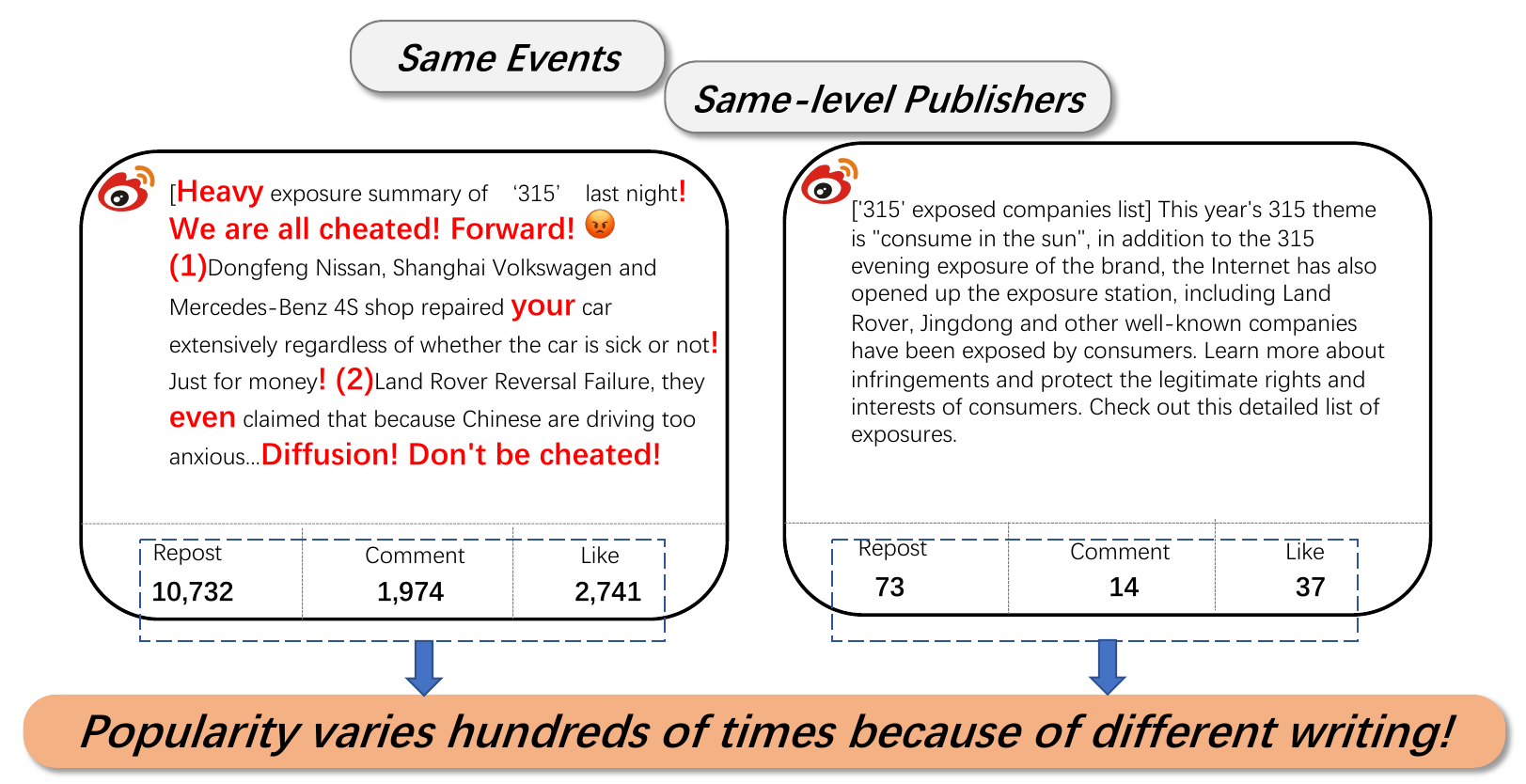}
	\caption{Examples to show the relationship of writing style and quality(popularity). Two pieces of news are reported by CCTV News and Xinhua Viewpoint respectively (all belong to very authoritative news media with more than 50 million followers). They reported the same event with great difference in description. Post1 (the left one) receives higher popularity compared with Post2 (the right one). It tends to use `we, your' to show that event reported is related users and interact with them. It also uses many `!' to attract users' attention and `(1), (2), (3)...' to make news more readable. This paper focuses on mining such writing style and analyzes its relationship with quality.}
	\label{case_intro2}
\end{figure*}

Due to the rapid development of Internet technologies, online social networks are becoming popular. People are more and more willing to receive news on social networks. This social trend promotes traditional newspapers to report news on social networks. We observed that writing style can make a great difference on news quality: The popularity (a kind of measure for quality to some extent) of news which are posted by same-level publishers and describe a same event even varies hundreds of times due to different writing style as Figure \ref{case_intro2} shows.
Thus a question naturally arises: whether can we analyze the influence of writing style on news quality automatically and assist writers in improving news quality?

Generally, it is difficult to estimate `quality'  of text in a certain field (such as science articles~\cite{pitler2008revisiting}) without human judgement as the ground truth. However, for content on social networks, `popularity' can be seen as the natural ground truth for 'quality'. Methods that estimate information quality of web pages based on the number of their incoming links, which can be seen as a kind of measure for popularity, has been successful~\cite{kleinberg1999authoritative,page1999pagerank}. These facts implied that there is high correlation between the popularity and the quality of information. 

 However, simply treating `popularity' and `quality' as synonyms  is not reasonable  in our work. Since apart from writing style, popularity is also greatly influenced by event and publisher. We design methods to weaken their influence. In this context, we can regard `popularity'  as the measure for `quality'. 

News on social networks describe events in concise and accessible language and wish to attract user's attention to participate in them. To achieve these goals, some universal writing guidelines are followed by news writers like `news should be written interestingly', which are introduced to in this paper.
  
Previous works often focused on predicting quality precisely but gave little insights into how to write. Our work steps further to analyze why the quality score is given and present some accessible suggestions on how to improve it. 

Our contributions are summarized as follows.

\begin{enumerate}
\item We propose eight types of linguistic features  based on eight news writing guidelines to mining writing style which may influences news quality on social networks. 
\item We analyze the relationship of these features and news quality from both Inter-User and Intra-User. From Inter-User, as events reported by different users have great overlap, we can weaken its influence on popularity by analyzing among users. From Intra-User, user's influence including the number of followers can be alleviated within each user.
\item  Then we propose a social network news quality assessment model (SNQAM). SNQAM predicts the quality of news by weighted summation of all features according to their correlation with quality. It can also predict scores for the eight feature types respectively and give targeted suggestions according to the corresponding guideline. 

\end{enumerate}

The experimental results show that the style features indeed have high correlation with news quality and prediction results for quality are reasonable  and interpretable.

\section{Related Work}
Works about assessment of information quality have made great progress for many types of information. \cite{DBLP:journals/corr/abs-1905-08949} focused on the assessment of neural question generation. \cite{lei-etal-2019-revisit} proposed an effective machine translation (MT) evaluation. \cite{kleinberg1999authoritative} estimated quality of web pages based on the number
of their incoming links; ~\cite{gu2015using} proposed a new no-reference image quality assessment metric using the recently revealed free-energy-based brain theory and classical human visual system-inspired features. ~\cite{pitler2008revisiting,louis2013makes} take into account various linguistic factors to produce predictive models for article quality. Few works focused on news on social networks.

It is generally difficult to estimate `quality' without human intervention especially for articles. Pilter et al. used human judgement as the ground truth of article quality while Louis et al. classified articles which were appeared in “The Best American Science Writing” as high-quality articles. To solve this problem, popularity-based methods have been widely used. These works often focus on improving prediction accuracy and pay more attention on factors besides writing style including publisher's social context, information diffusion model, etc. ~\cite{cui2011should} analyze the relationship between user's information and content popularity; ~\cite{DBLP:conf/ijcai/WuCZHLM17,DBLP:conf/www/GaoSLC16}  used temporal information and model the diffusion of information to predict popularity. All of them focus on predicting popularity but can give little insights on how to write.

\section{Facets of news writing}
We discuss eight prominent facets of news writing which we hypothesized will have an impact on news quality. Some facets such as Readability and Formality are used widely in the tasks of text analysis like clickbait detection~\cite{DBLP:conf/aaai/BiyaniTB16} and authorship attribution~\cite{gu2015using}. Others are proposed by us according to some news writing guidelines. For each facet, we first define some basic linguistic features then combine them to get a high-level feature.
All features are defined within a piece of news and summarized in Table \ref{features}. We get 45 basic features and 8 high-level features in all.

\begin{table*}[t]
\centering
\begin{tabular}{p{2cm}p{6cm}|p{2.5cm}p{6cm}}  
\toprule
Facets  & Features &Facets  & Features\\
\midrule
Readability&Sentence\_broken, Characters, Words, Sentences, Clauses, Average\_word\_length, Professional\_words, RIX, LIX, LW& Logic&Forward\_reference, Conj \\
\midrule
Credibility&@, Numerals, Official\_speech, Time, Place, Object, Uncertainty, Image&Formality&Noun,  Adj, Prep, Pron, Verb, Adv, Sentence\_broken\\
\midrule
Interactivity&Question\_mark, First\_pron, Second\_pron, Interrogative\_pron&Interestingness&Rhetoric, Exclamation\_mark, Face, Idiom, Adversative, Adj, Image\\
\midrule
Sensation&Sentiment\_score, Adv\_of\_degree, Modal\_particle, First\_pron, Second\_pron, Exclamation\_mark, Question\_mark&Integrity&HasHead, HasImage, HasVideo,  HasTag, HasAt, HasUrl\\

\bottomrule
\end{tabular}
\caption{Linguistic features (some features can be classified into different facets).}
\label{features}
\end{table*}

\paragraph{Readability: Being clear and easy to read is a basic requirement of news.}Especially for social network like Weibo, the length of posts is limited to 140 words and users generally spend less time to quickly browse the news. 

Sentence\_broken~\cite{xiao2015a}, Characters, Words, Average\_word\_length, Sentences, Clauses~\cite{pitler2008revisiting}, RIX, LIX and LW~\cite{anderson1983lix} are all proposed before to capture Readability in a piece of news. In addition, considering that users of social network are involved in all walks of life, too many professional terms will make users confusing. Therefore, we propose Professional\_words to count their number.
The smaller the above 10 features, the higher the readability. Based on them, 

 {\it Readability = -(Sentence\_broken + Characters+ Words + Sentences + Clauses + Average\_word\_length + Professional\_words+ LW +  RIX + LIX)}

\paragraph{Logic: Good news should be logical and contextually coherent.} Forward\_reference~\cite{DBLP:conf/aaai/BiyaniTB16} is used to capture logic of news which can create a sort of tease or information gap between several sentences. It includes demonstratives (this, that, those, these...) and third person pronouns (he, she, his, her, him). Conjunctions are also used to make context more coherent, like `so, and'. Thus, {\it Logic = Forward\_reference + Conjs}

\paragraph{Credibility: News should be rigorous and credible, especially for official media.}

For news media In Weibo, `@' is often used to bring out the object of the event. Sometimes it is used to elicit the source of the news. Detailed numbers and relevant images can make the news more authentic. Official\_speech counts the number of words which indicate that information is released by the official institutions, such as `Circulated'. Time, Place, Object respectively correspond to the three elements of news people often said: when, where and who. It's important for news to conclude such elements. Uncertain words shouldn't be used as they will confuse readers including `maybe, perhaps, ...'. 

{\it Credibility = @ + Numerals + Official\_speech + Time +  Place + Object - Uncertainty + Image }

\paragraph{Formality: News on social network tend to be more spoken than custom newspaper.}
Referring to the formality definition in English~\cite{DBLP:conf/aaai/BiyaniTB16}, Formality of news is related to the number of various parts of speech including Noun,  Adj, Prep, Pron, Verb and Adv. Sentence\_broken is also considered as the higher formality the text, the less pauses in the sentence~\cite{gu2015using}. Thus, 

{\it Formality = Noun + Adj + Prep - Pron -  Verb - Adv - Sentence\_broken  }

\paragraph{Interactivity: News with higher interactivity can cause readers to think and promote them participate in the discussion.}  Sentences like "How about you?" can achieve this effect. So, we counts the number of question marks, first pronouns, second pronouns and interrogative pronouns in news. 

{\it Interactivity = Question\_mark + First\_pron + Second\_pron + Interrogative\_pron}

\paragraph{Interestingness: Naturally, interesting description can attract more readers}

Rhetoric counts the number of rhetorical devices used in news. For example,

    {\it Today, `Little Penguin' is 15 years old.}  
    
Because Tencent QQ, a popular chat software in China, has the icon like a penguin,  the news compared it to `Little Penguin'. Such description is interesting. Usually the content of rhetoric will be extracted with ', so Rhetoric is counted by the number of '. Similarly, Idiom, expressions, Exclamation\_mark and images are considered. Adversative words like `however, but' can enhance the drama of the story. Adjectives are used to modify nouns and enrich description~\cite{DBLP:conf/acl/McIntyreL09}. Based on above seven features, Interestingness of a piece of news is defined as,

{\it Interestingness = Rhetoric +  Exclamation\_mark + Face +  Idiom + Adversative + Adj + Image}
\paragraph{Sensation: Good news can impress people and  cause a sensation.} %add related work
Content with distinct emotional orientation is obviously easier to be noticed by readers in a huge and continuous flow of information. Emotional expression has become an important means for news content to gain attention.
We calculate sentiment score of a piece of news by emotional dictionary matching[here add where dictionary from]. The sentiment score is between -1 and 1, where negative value means negative emotion, positive value means positive emotion and the greater the absolute value, the stronger the emotion. Adv\_of\_degree means the number of degree adverbs, like `too, very'. Modal\_particle is the number of modals, like `Ah'. Similarly, Sensation has positive correlation with First\_pron, Second\_pron, Exclamation\_mark and Question\_mark as well. We define Sensation as follows,

{\it Sensation =Sentiment\_score + Adv\_of\_degree
	+ Modal\_particle +  First\_pron
	+   Second\_pron
	+   Exclamation\_mark
	+   Question\_mark}

\paragraph{Integrity: Some parts are essential for news like title.}
We observed that the news media on social networks generally have the following basic structure: title, image, video, topic(\#), @ and web link. Since the title is the focus of catching the reader’s eye, tags (\#) can help news pushed to more people, and People are more willing to read multimodal information, we give these features higher weights when calculating the Integrity of a piece of news,.

{\it Integrity = 2*HasHead + 2*HasImage 
	+ 2*HasVideo+  2*HasTag +  HasAt
	 +  HasUrl}

Most of the features can be obtained directly by HanLP\footnote{\url{https://github.com/hankcs/HanLP}}, an open source Chinese natural language processing package including functions like word segmentation, part-of-speech tagging, syntax analysis, etc.  Its part-of-speech category is abundant. In addition to the basic part-of-speech like nouns, it can also recognize time words, name, various symbols, etc, which can be used to obtain Time, Object and others. Other features, such as Adversative and Adv\_of\_degree, were obtained by querying the Chinese dictionary and performing dictionary matching.

\section{Corpus and Quality Measure }
\subsection{Corpus}
\subsubsection{Data Collection}
We chose Weibo as the research platform, a Chinese social network similar to Twitter with more than 337 million users by June 2018. Most traditional news media create their accounts on Weibo for publishing news and interacting with people. 

This paper focus on news media. We chose 4 authoritative news accounts on Weibo according to the list published by {\it People's Network Opinion Center\footnote{\url{http://yuqing.people.com.cn/n1/2018/1214/c364056-30467317.html}}} (which analyzed the media's
readings, forwarding numbers and comments and evaluated their influence on social networks). They are {\it People’s Daily}, {\it CCTV News}, {\it Xinhua Net} and {\it Xinhua Viewpoint}. All of them locate at top 30 of the list and belong to influential accounts. We collected all of their posts since registered. 

\subsubsection{Data Preprocessing}
It has been previously suggested that 85\% of posts received 80\% of reposts within 48 hours on Weibo ~\cite{ma2013towards}. So, our data set retained posts published for more than a week; Due to the different time of the first post, we have kept posts between July 2012 and November 2018; On Weibo, there exits sweepstakes. Such posts often require users to forward them and draw prizes, so the popularity will be higher. They are noises for our analysis, so we unify all users'  lottery posts; As the popularity of forwarding post cannot be distinguished whether resulted from this post or the source. Therefore, we only keep original posts.
Finally, we get data size in Table \ref{followers}.

\begin{table}[h]
	\centering
	\begin{tabular}{lccc}  
		\toprule
		Accounts & Followers & Posts & Quality \\
		& (million) & &  \\
		\midrule
		People's Daily & 8,268 & 82,788 & 8647.1\\
		CCTV News & 7,789 & 84,583 & 6048.9\\
		Xinhua Net & 5,115 & 67,765 & 471.4\\
		Xinhua Viewpoint & 5,180 & 72,152 & 506.0\\
		\bottomrule
	\end{tabular}
	\caption{ `Quality' here refers to average sum of `like', `comment' and `repost' for each piece of news.}
	\label{followers}
\end{table}\textit{}
Their average qualities on Weibo are quite different as Table \ref{followers} shows. So we guessed that there are some differences in their writing style, which lead to the difference of quality. 

\subsection{Quality Measure}
Quality of news on Weibo can be measured by popularity easily, including `repost', `like' and `comment'. We define quality as the sum of these indicators as they have strong correlation with each other (Spearman Rank Correlation, SRC,  is larger than 0.75). 

\section{Methods and Experiments}

We analyze the usefulness of the eight types of linguistic features for news quality assessment from both Inter-User and Intra-User. Then based on them, we propose a news quality assessment model.

\subsection{Feature Analysis: Inter-User}
\begin{figure}[h]
	\centering
	\includegraphics[scale=0.4]{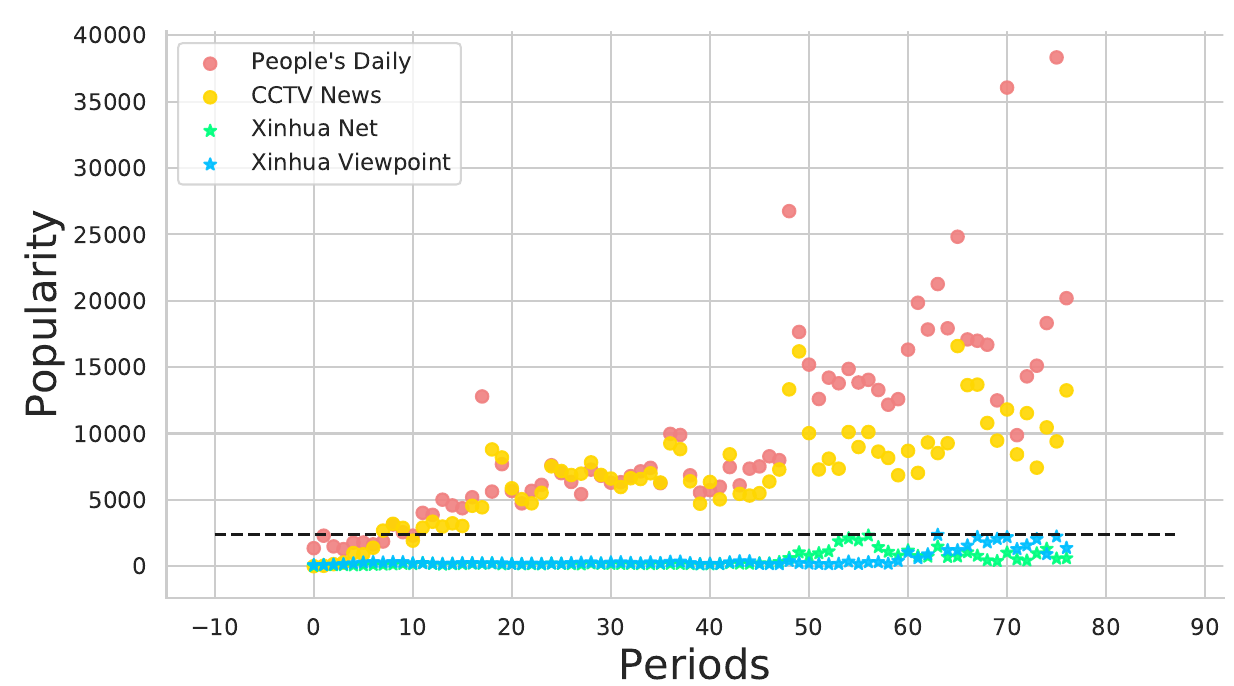}
	\caption{Quality of People's Daily and CCTV News are always dozens of times higher than the others (Data is too large to observe intuitively, thus we calculate average popularity of 4 accounts per period (30 days) from July 2012 to November 2018).}
	\label{popularity_differ}
\end{figure}

From Inter-User, as events reported by different users have great overlap, we can weaken its influence on popularity by analyzing among users.
Although the four media we selected all belong to influential media on Weibo, their qualities are quite different and can be divided into two levels. Most qualities of posts on People’s Daily and CCTV News are higher than that of the remaining two media and differs greatly as Figure \ref{popularity_differ} shows. Therefore, we classified all the posts of the high-quality accounts (People’s Daily and CCTV News) as VERY GOOD and all the posts of typical accounts ( Xinhua Net and Xinhua Viewpoint) as TYPICAL. By comparing the differences in writing style between these two types of accounts, we can find out the reasons for making these accounts high-quality. 

\begin{figure}[h]
\centering
\includegraphics[angle=270,width=0.45\textwidth]{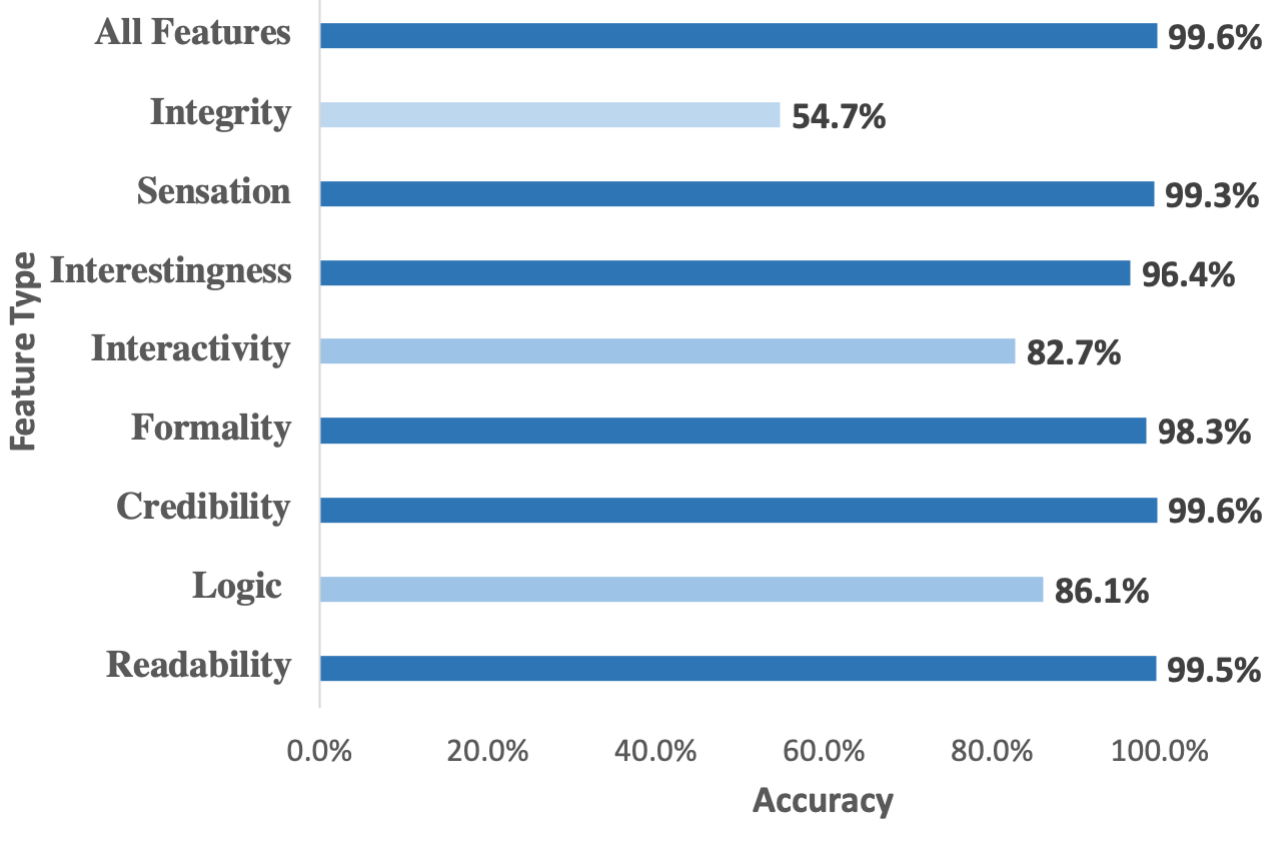}
\caption{Average accuracy of test in 5-fold cross validation for various feature types.}
\label{class_result}
\end{figure}

\paragraph{Classification between VERY GOOD and TYPICAL:}We chose Random Forest (RF) as the classifier and optimized the model on the dataset using 5-fold cross validation. We divided the dataset into 5 parts, train on 4 parts and test on the held-out data. Experiments based on all features and each type of news quality features respectively with same RF parameters are performed. The classification results about the average accuracy of 5 experiments obtained are shown in Figure \ref{class_result}.

\begin{figure}[h]
	\centering
	% scale=0.35
	\includegraphics[width=0.5\textwidth]{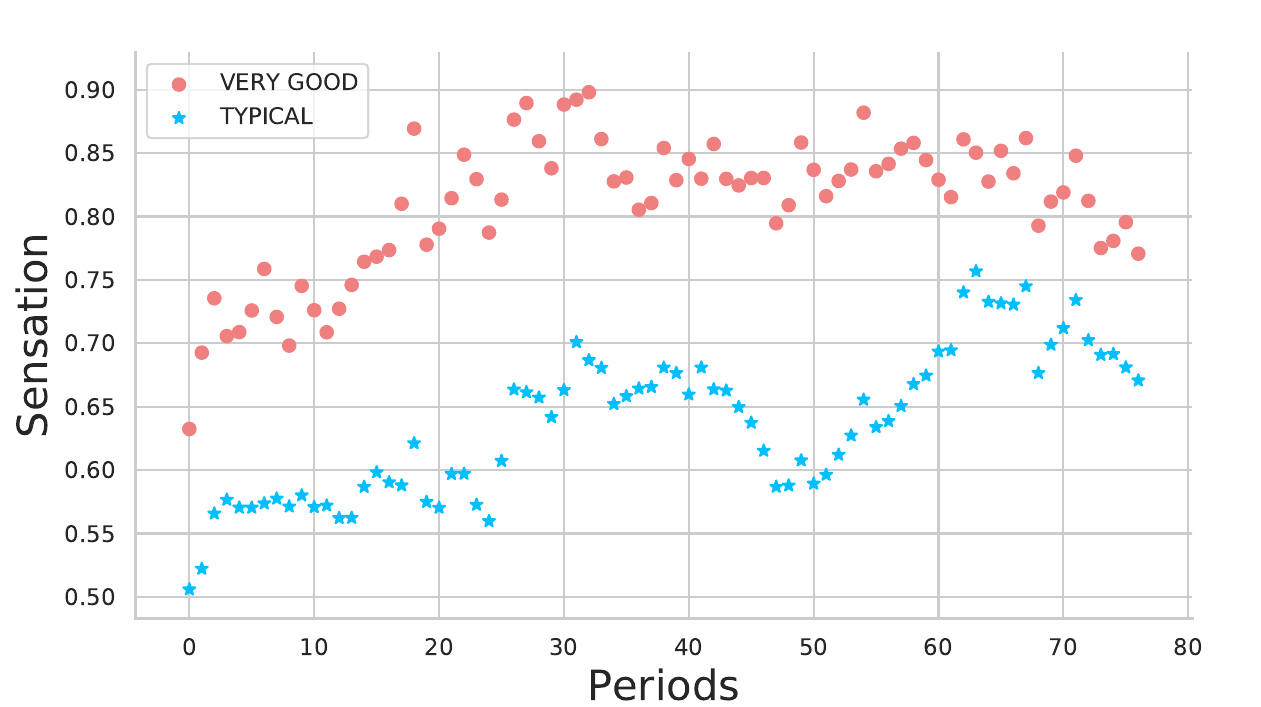}
	\caption{Sensation of VERY GOOD tend to be higher than that of TYPICAL.}
	\label{sensation}
\end{figure}

As Figure \ref{class_result} shows,  classification with all features achieve \textbf{99.6\%} accuracy which implies that writing style have great difference on VERY GOOD  and TYPICAL news. Moreover, to analyze the performances of each feature type, we performed same experiments with only one type respectively. Experimental results show that only using \textbf{Readability}, \textbf{Credibility}, \textbf{Formality}, \textbf{Interestingness} or \textbf{Sensation} features all performed well with accuracy more than \textbf{96\%}. It indicates that the corresponding five writing guidelines are very helpful in news writing. Take Sensation as an example, as shown in Figure \ref{sensation}, news of VERY GOOD are always more emotional, which encourages news writers to use stronger emotional expression when describing events to arouse people's emotional resonance. 

Remaining style feature type including Logic and Interactivity give accuracy 86.1\% and 82.7\% respectively,  implying moderate correlation with news quality. Integrity only gives 54.7\% accuracy mainly resulting from that posts in our data set are all published by authoritative media and in standard format. Therefore, we can nearly ignore it when evaluating quality for news.

\paragraph{Feature Importances:}Importances of features for RF classification with all features are presented in Table \ref{feature_importance}. 
\begin{table}[h]
	\centering
	\begin{tabular}{ll}  
		\toprule
		Rank  & Feature \\
		\midrule
		1  & \ce{Credibility^C}    \\
		2  & \ce{Average\_word\_length^R}     \\
		3  & \ce{Characters^R}     \\
		4  & \ce{Exclamation\_mark^{IE,S}}     \\
		5  & \ce{Sentences^R}    \\
		6  & \ce{Words^R}     \\
		7  & \ce{Professional\_words^C}      \\
		8  & \ce{Sentiment\_score^S}     \\
		9  & \ce{Numerals^C}      \\
		10  & \ce{Rhetoric^{IE}}    \\
		\bottomrule
	\end{tabular}
	\caption{Top 10 important features in classification. Upper corner means corresponding feature type, such as \ce{Average\_word\_length^R} means feature Average\_word\_length belongs to Readability (R). Some features belong to more than one category. }
	\label{feature_importance}
\end{table}\textit{}

Top 10 important features are mostly belong to \textbf{Readability}, \textbf{Credibility}, \textbf{Interestingness} and \textbf{Sensation}, which are also the types performed individually with accuracy more than 96\% mentioned before. It further confirms the importance of these types of style in news quality assessment and gives us insights into where we should improve more specifically. Like Sensation mentioned before, news writes can improve news quality by improve Sensation mainly focusing on using more exclamatory marks and emotional words.

\subsection{Feature Analysis: Intra-User}
Within each user, user's influence on popularity including the number of followers can be alleviated. By mining common laws for all users, we can analyze the relationship of writing style and news quality.

\paragraph{Correlation between Linguistic Features and Quality:}
We calculate the SRC between feature and quality for each user and present 10 common and most influential features for all users in Table \ref{corr3} .

\begin{table}[h]
	\centering
	\begin{tabular}{lcc}
		\toprule
		\multirow{2}{*}{Feature} & \multicolumn{2}{c}{Correlation} \\
		\cmidrule(lr){2-3}
		& mean & standard deviation \\
		\midrule
		\ce{Interestingness^{IE}}	&0.347 &	0.028 
 \\
		\ce{Exclamation\_mark^{IE,S}} &	0.306 &	0.048 
 \\
		\ce{Image^{IE, C}} &	0.303 &	0.060 
 \\
		\ce{Average\_word\_length^R} &	-0.242 &	0.030 
 \\
		\ce{Sensation^S}&	0.234 &	0.066 
 \\
		\ce{LIX^R}	&-0.226 &	0.018 
 \\
		\ce{RIX^R}	&-0.217 &	0.013 
 \\
		\ce{Pron^F}&	0.214 &	0.054 
 \\
		\ce{Second\_pron^{IR,S}} &	0.192 &	0.043 
 \\
		\ce{Forward\_reference^L}&	0.186 &	0.050  \\
		\midrule
	\end{tabular}
	\caption{Mean and standard deviation are calculated on the correlation of all users. The greater the absolute value of the mean, the greater the correlation. Positive value means positive correlation.}
	\label{corr3}
\end{table}
We find that the influences of features have great overlap among all users. First, the standard deviation is relatively small which means the correlation of features and quality are very similar within each user. Second, there exits 10 common features among top 20 correlated features  of all users. It indicates, for each user, features playing most important roles are similar as well. 

Moreover, compared features in Table \ref{corr3} and Table \ref{feature_importance}, we find that the main feature types they belong to are similar: \textbf{Readability}, \textbf{Credibility}, \textbf{Interestingness} and \textbf{Sensation}, which proves that no matter within users or among users, these writing guidelines are most important and should be paid more attention. 

\begin{figure}[h]
	\centering
	\includegraphics[width=0.4\textwidth]{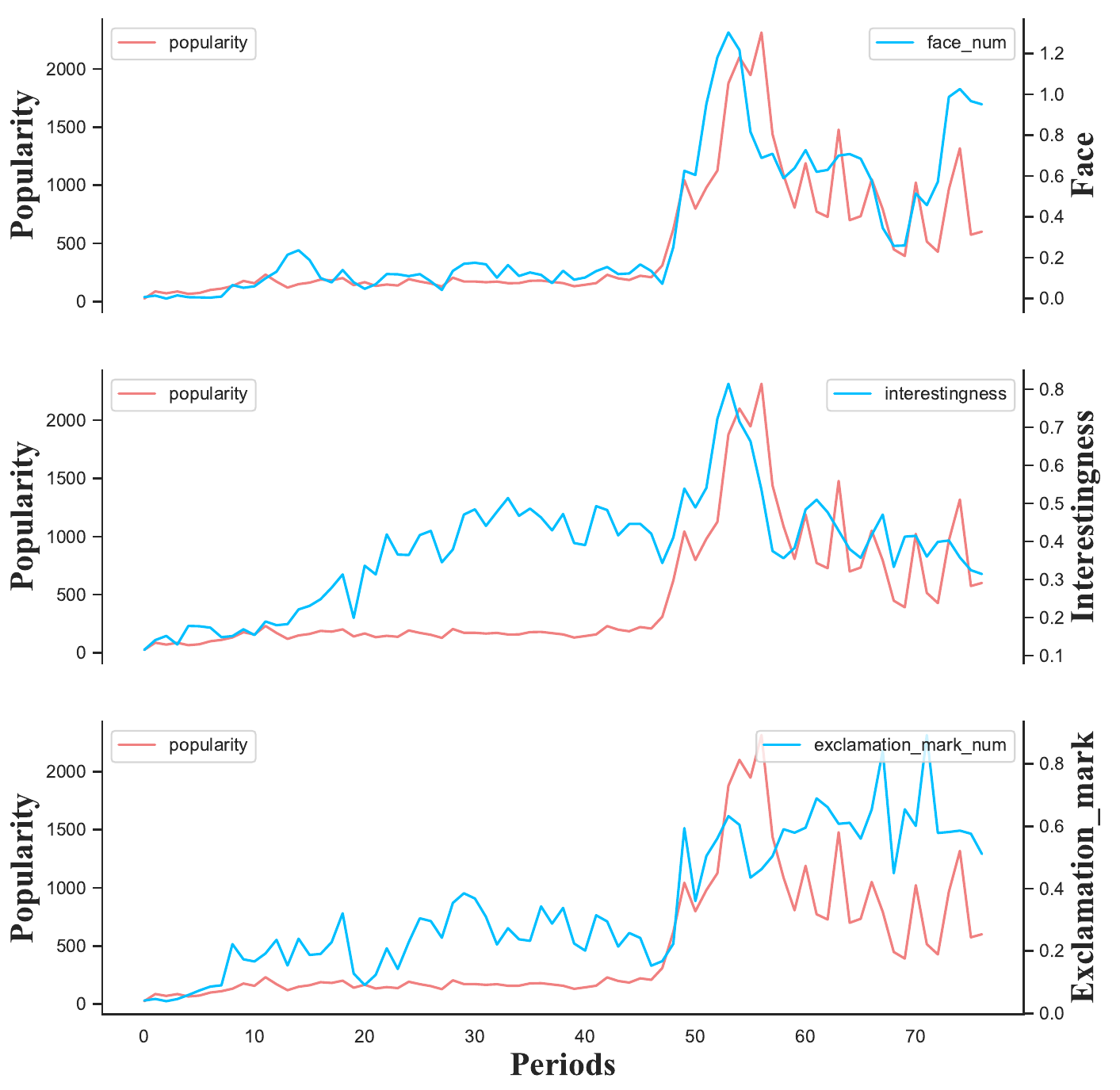}
	\caption{ Three features which have most similar change to quality (\textbf{Face}, \textbf{ Interestingness} and \textbf{Exclamation\_mark}).}
	\label{XinhuaNet_pop_fea}
\end{figure}

\paragraph{Quality Drift:} According to researches on authorship attribution research~\cite{azarbonyad2015time}, there exits writing style drift when time span is long. Observing the user's quality trend curves over time, we find a interesting phenomenon: For Xinhua Net, its quality has suddenly increased after 50th periods and then keep a stable and little higher level than before 50th periods. To figure out if such change is related with writing style, we compare all features time trends and find that some are very similar to quality trend as Figure \ref{XinhuaNet_pop_fea} shows. 

After 50th periods, Xinhua Net began to use more faces, exclamation marks and describe news more interesting which are all belong to the most important feature types for quality (\textbf{Interestingness} and \textbf{Sensation}). It was a wise reform and received good reward which further implies that by changing writing style analyzed in this paper, news quality can be improved in application.

\subsection{News Quality Assessment Model}
Although in Section Intra-User, experiments for classifying news into VERY GOOD and TYPICAL performed even 99\% accuracy based on writing style features, such classification seems not very appealing in reality. To give accessible suggestions on improving quality, besides classification, we also want to give a interpretable quality score. Therefore, based on analysis above, we propose the following news quality assessment model.
\begin{align}
News\_Quality\_Score = \sum_{i=1}^{n} W_i*F_i \\
W_i = I_i * C_i
\end{align}
where n refers to the number of features, $W_i$ is the weight of each feature for news quality assessment and $F_i$ is the value of $i_{th}$ feature. $W_i$ is calculated by multiplying feature importance $I_i$ (when classifying posts into VERY GOOD and TYPICAL in Intra-User) and correlation $C_i$ between features and news quality mentioned in Inter-User (as the correlation of features in each user are slightly different, we use the correlation calculated in all user's posts.).

The quality assessed by the above simple model performs strong relationship with the corresponding quality obtained on social networks (popularity), which achieves 0.606 SRC. 

\begin{figure}[h]
	\centering
	%angle=270,
	\includegraphics[width=0.4\textwidth]{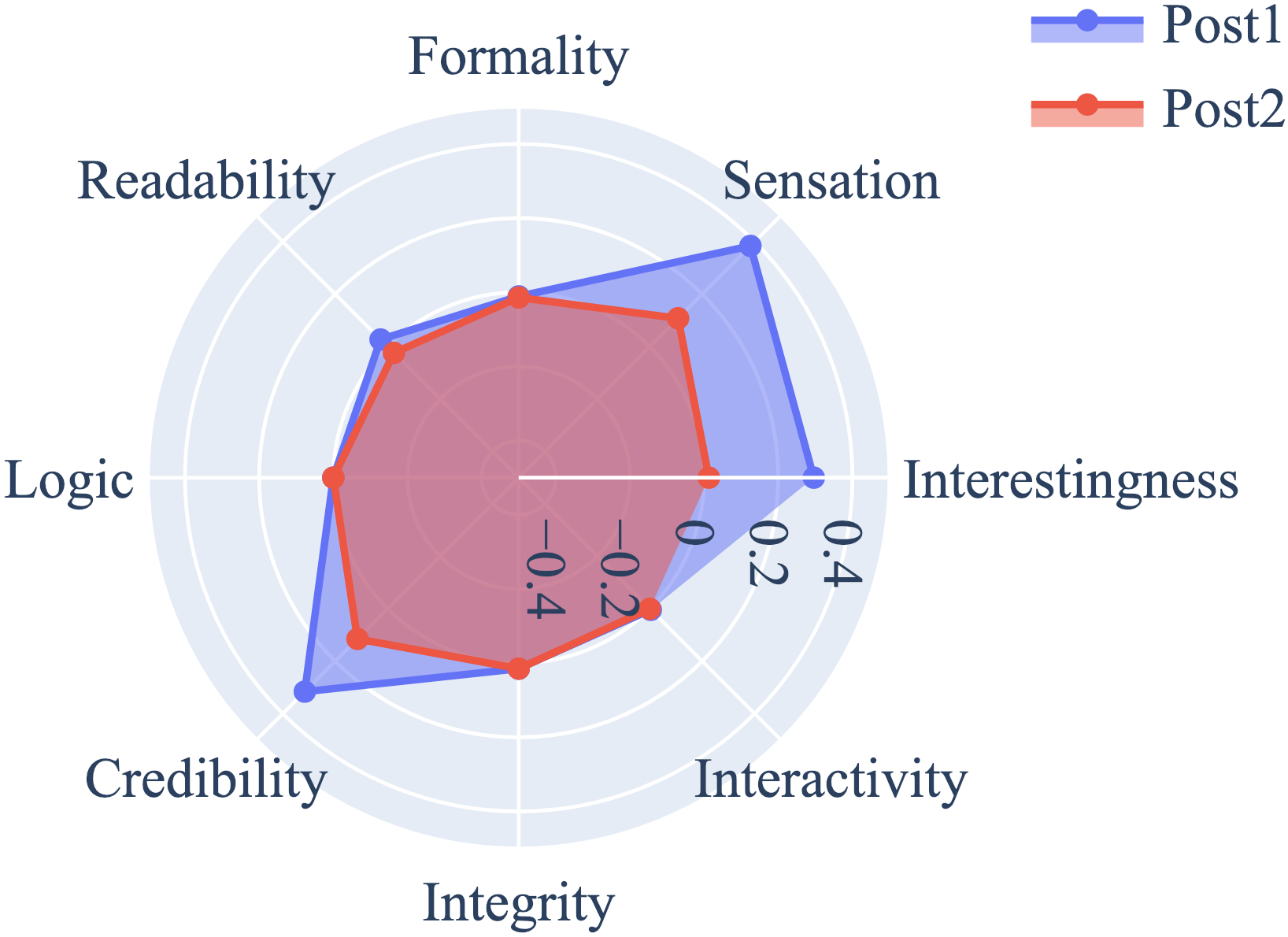}
	\caption{Compared with Post1, Post2 has low quality mainly due to its bad performance on \textbf{Sensation (0.384 vs. 0.108)}, \textbf{Interestingness (0.296 vs. 0.013)} and \textbf{Credibility (0.316 vs. 0.115}).}
	\label{quality_analysis}
\end{figure}

In addition, the score can be seen as calculating by summing the scores of eight news writing facets. So, by analyzing what facet gets low score, we can give the targeted suggestions following by the corresponding writing guidelines. Take an example, for two posts introduced in Introduction, our model predict their quality with 0.770 (Post1) and 0.207 (Post2) respectively. Figure \ref{quality_analysis} shows the detailed analysis for eight writing facets.

Thus we suggest that Post1 can be improved by focusing on improving the three feature types such as using some faces to stimulate the reader's emotions, more digits and detailed description to improve Credibility, and describing events in more interesting and appealing ways like using `!' as Post1 did.

\section{Conclusion}
In this paper, we analyze the influence of writing style on news quality. Firstly, we propose eight types of linguistic features based on eight news writing guidelines to mining writing style which may influences news quality and analyze the relationship of these features and news quality from both Inter-User and Intra-User. Then we propose a simple but interpretable model to predict the quality of news and give targeted suggestions on how to improve quality. The experimental results demonstrate the efficacy of these features.

In future work, we will try to explore news generation
techniques to generate high-quality news followed by these writing guidelines.

%% The file named.bst is a bibliography style file for BibTeX 0.99c
\bibliographystyle{named}
\bibliography{howtowrite}

\end{CJK}  

\end{document}